\documentclass[10pt,twocolumn,letterpaper]{article}

\usepackage{iccv}
\usepackage{times}
\usepackage{epsfig}
\usepackage{graphicx}
\usepackage{amsmath}
\usepackage{amssymb}
\usepackage{tabu}
\usepackage{color}
\usepackage{bm}
\usepackage[font=small]{caption}
\usepackage{subcaption}
\DeclareCaptionSubType{subfigure}

\AtBeginDocument{
  \let\oldsubsubfigure=\subsubfigure
  \renewcommand{\subsubfigure}{\expandafter\def\csname @captype\endcsname{subfigure}%
    \oldsubsubfigure}%
}

\newcommand{\rulesep}{\unskip\ \vrule\ }

\usepackage[pagebackref=true,breaklinks=true,letterpaper=true,colorlinks,bookmarks=false]{hyperref}

\iccvfinalcopy 

\def\iccvPaperID{4301} 

\ificcvfinal\fi

\def\cD{\mathcal{D}}

\begin{document}

\title{VDM-DA: Virtual Domain Modeling for Source Data-free Domain Adaptation}

\author
{Jiayi Tian\thanks{means equal contribution} \textsuperscript{1}, Jing Zhang\footnotemark[1] \textsuperscript{1}, Wen Li\textsuperscript{2}, 
Dong Xu\textsuperscript{3}\\
\textsuperscript{1}Beihang University , 
\textsuperscript{2}University of Electronic Science and Technology of China \\
\textsuperscript{3}University of Sydney \\

{\tt\small tianjy@buaa.edu.cn, zhang\_jing@buaa.edu.cn, liwenbnu@gmail.com, dong.xu@sydney.edu.au }
\vspace{-10pt}

}

\maketitle
\ificcvfinal\thispagestyle{empty}\fi

\vspace{-10pt}
\begin{abstract}
\vspace{-10pt}
Domain adaptation aims to leverage a label-rich domain (the source domain) to help model learning in a label-scarce domain (the target domain). Most domain adaptation methods require the co-existence of source and target domain samples to reduce the distribution mismatch, however, access to the source domain samples may not always be feasible in the real world applications due to different problems (\eg, storage, transmission, and privacy issues). In this work, we deal with the source data-free unsupervised domain adaptation problem, and propose a novel approach referred to as Virtual Domain Modeling (VDM-DA). The virtual domain acts as a bridge between the source and target domains. On one hand, we generate virtual domain samples based on an approximated Gaussian Mixture Model (GMM) in the feature space with the pre-trained source model, such that the virtual domain maintains a similar distribution with the source domain without accessing to the original source data. On the other hand, we also design an effective distribution alignment method to reduce the distribution divergence between the virtual domain and the target domain by gradually improving the compactness of the target domain distribution through model learning. In this way, we successfully achieve the goal of distribution alignment between the source and target domains by training deep networks without accessing to the source domain data. We conduct extensive experiments on benchmark datasets for both 2D image-based and 3D point cloud-based cross-domain object recognition tasks, where the proposed method referred to Domain Adaptation with Virtual Domain Modeling (VDM-DA) achieves the state-of-the-art performances on all datasets. 

\end{abstract}

\vspace{-15pt}
\section{Introduction}
\label{sec:intro}
Deep neural networks (DNNs) have achieved remarkable performance in a wide range of computer vision tasks. However, the superior performance of the data-driven DNNs methods highly relies on a large amount of well annotated data. 
To reduce the efforts and costs from the labor-intensive data annotation task, 
a popular solution is to transfer knowledge from a related source domain with abundantly labeled data to a domain of interest (\ie the target domain) with minimal or even no labeled training data. To alleviate the performance drop caused by data distribution mismatch between the source domain and the target domain, many unsupervised domain adaptation (UDA) methods have been proposed to adapt a model to an unlabeled target domain from a labeled source domain~\cite{Pan2009,Duan2012,Long2013,Sun2016,Long2015,Long2017,Ganin2016,Bousmalis2017, Hoffman2018,Saito2018,Zhang2018,Zhang2020, kang2019contrastive, tang2020unsupervised, 5288526}.

Most domain adaptation methods require both source and target samples for learning the domain adaptation models. However, in practice, the requirements of accessing to the source domain data can be very restrictive in the real world applications. For example, the source domain data are generally assumed to be in a large scale to cover more transferable knowledge, and are thus difficult to be stored, transmitted, and processed. Moreover, sharing data is also becoming a concern due to privacy or security issues. This leads to the more challenging source data-free UDA (SFUDA) problem, \ie, given the pre-trained source model and unlabeled target domain samples, how to improve the classification performance in the target domain without accessing to the original labeled data in the source domain.

In recent years, several SFUDA methods have been proposed~\cite{Liang2020,Kundu2020,Li2020}. These methods either borrow the idea of self-training~\cite{Liang2020,Kundu2020} or generate the source domain images based on the pre-trained source model~\cite{Li2020}. The data generation module in~\cite{Li2020} is very complicated and cannot generalize well to more complex input data, while the self-training methods~\cite{Liang2020,Kundu2020} are generally self-referential and may suffer from error aggregation.

In this work, we propose a new SFUDA approach through virtual domain modeling (VDM). We aim to address the core issue in SFUDA, \textbf{\emph{how can we reduce data distribution mismatch between the source and target domains without accessing to the original source samples}}? To this end, we introduce an intermediate virtual domain in the high-level compact feature space to bridge the unseen source data distribution and the target domain distribution. Consequently, we convert the problem of reducing data distribution mismatch between the source and target domains as minimizing the domain gap between source and virtual domains as well as that between the virtual and target domains. 

To archive this goal, we consider two aspects. On one hand, we generate the virtual domain samples by using an approximated Gaussian Mixture Model (GMM) in the feature space with the aid of the pre-trained source model, such that the virtual domain can still maintain a similar distribution with the source domain without accessing to original source data. On the other hand, we also propose an effective distribution alignment method to reduce the distribution gap between the virtual domain and the target domain by gradually improving the compactness of the target domain distribution through model learning. In this way, we successfully achieve the goal of distribution alignment between the source and target domains by training deep networks without accessing to the source domain data. Our experimental results clearly demonstrate the effectiveness of the proposed approach.

The main contributions of this paper are summarized as follows,
\begin{itemize}
    \item We propose a new generic source data-free unsupervised domain adaptation method, named Domain Adaptation with Virtual Domain Modeling (VDM-DA), by modeling a compact intermediate virtual domain in the feature space, which acts as a bridge between the source and target domains for explicitly reducing the data distribution mismatch.

    \item The virtual domain is constructed based on a simple and effective Gaussian Mixture Model (GMM) in which the unknown parameters are approximated by discovering the information hidden in the source model without introducing any extra learnable parameters. In addition, the unlabeled data in the target domain are then aligned with the constructed compact virtual domain by using our newly proposed uncertainty guided alignment approach.
    \item Our method is generic and can be readily used in various cross-domain applications, such as image-based and point-cloud-based object recognition. Extensive experiments for both 2D and 3D tasks demonstrate the newly proposed VDM-DA method outperforms the state-of-the-art UDA and SFUDA methods. 
\end{itemize}

\section{Related Works}
\noindent\textbf{Unsupervised Domain Adaptation} 
In vanilla unsupervised domain adaptation (UDA), the most prevailing paradigm is learning domain-invariant representations by minimizing the discrepancy between the source and target domain distributions. Two commonly used approaches are statistical moments-based methods~\cite{Pan2009,Duan2012,Long2013,Sun2016,Long2015,Long2017,sun2016return} and adversarial learning-based methods\cite{Ganin2016,Bousmalis2017, Hoffman2018,Saito2018, Zhang2018, Zhang2020,volpi2018adversarial,liu2020open,saito2018open,cao2018partial,you2019universal} that are motivated by Generative Adversarial Nets (GAN)~\cite{Goodfellow2014}. Both lines of works can directly reduce the distribution discrepancy without requiring explicit density estimation. The statistical moments-based methods mostly reduce the Maximum Mean Discrepancy (MMD) between the source and target distributions with carefully designed kernel functions in order to implicitly consider higher-order moments of data distribution. In contrast, the adversarial learning-based methods implicitly reduce the Jensen-Shannon divergence between different domains~\cite{Goodfellow2014}. In addition, some methods~\cite{Zhang2017,Zhang2018,Pan2019,Zhang2020,tang2020unsupervised,shu2018dirt} also borrow the idea of self-training to further exploit the target domain information. The most prevailing strategy is pseudo-labeling, where the highly confident unlabeled target samples are selected and then assigned with pseudo labels to further assist the training of the target model. 

The UDA approaches are also applied to different visual recognition tasks beyond image-based object recognition~\cite{chen2018domain,Qin2019,Hoffman2018,Bousmalis2017},.
For example, the PointDAN~\cite{Qin2019} method proposed a 3D-point-based UDA method by aligning the distributions of 3D objects across different domains both locally and globally. However, all the above methods require the original source domain data for effective domain adaptation.

\noindent\textbf{Source Data-free Unsupervised Domain Adaptation} 
By considering the practical problems in the real-world applications (\ie storage, transmission, and privacy issues), several recent works~\cite{Liang2020,Kundu2020,Li2020} studied the UDA problem without requiring the original source domain data. They either reconstructed the original source data~\cite{Li2020} or employed a self-training strategy with the entropy-based constraints~\cite{Liang2020,Kundu2020,Yang2020} and the pseudo-labeling technologies~\cite{Liang2020}. However, it is very difficult to reconstruct the source 
data in the original space which often requires complicated network architectures and advanced optimization technologies~\cite{Li2020}. In contrast, our method models a virtual domain in a much lower dimensional feature space without introducing additional learnable parameters. Moreover, the self-training strategy (i.e. the entropy minimization and the pseudo-labeling technologies) used in~\cite{Liang2020,Kundu2020,Yang2020} may suffer from error accumulation due to over-confident predictions produced by the source model. A recent arxiv work BAIT~\cite{Yang2020} introduces an additional classifier based on a similar motivation as minimax entropy~\cite{Saito2019}. Thus, the essential issue of reducing the distribution gap is not explicitly discussed in these works. In contrast, we mimic the distribution gap between the source and target domains through generating the virtual domain features, which provides both theoretical insights and superior empirical results for source data free unsupervised domain adaptation. 

\label{sec:method}
\begin{figure*}[t]
        \begin{center}
        \includegraphics[width=0.91\linewidth]{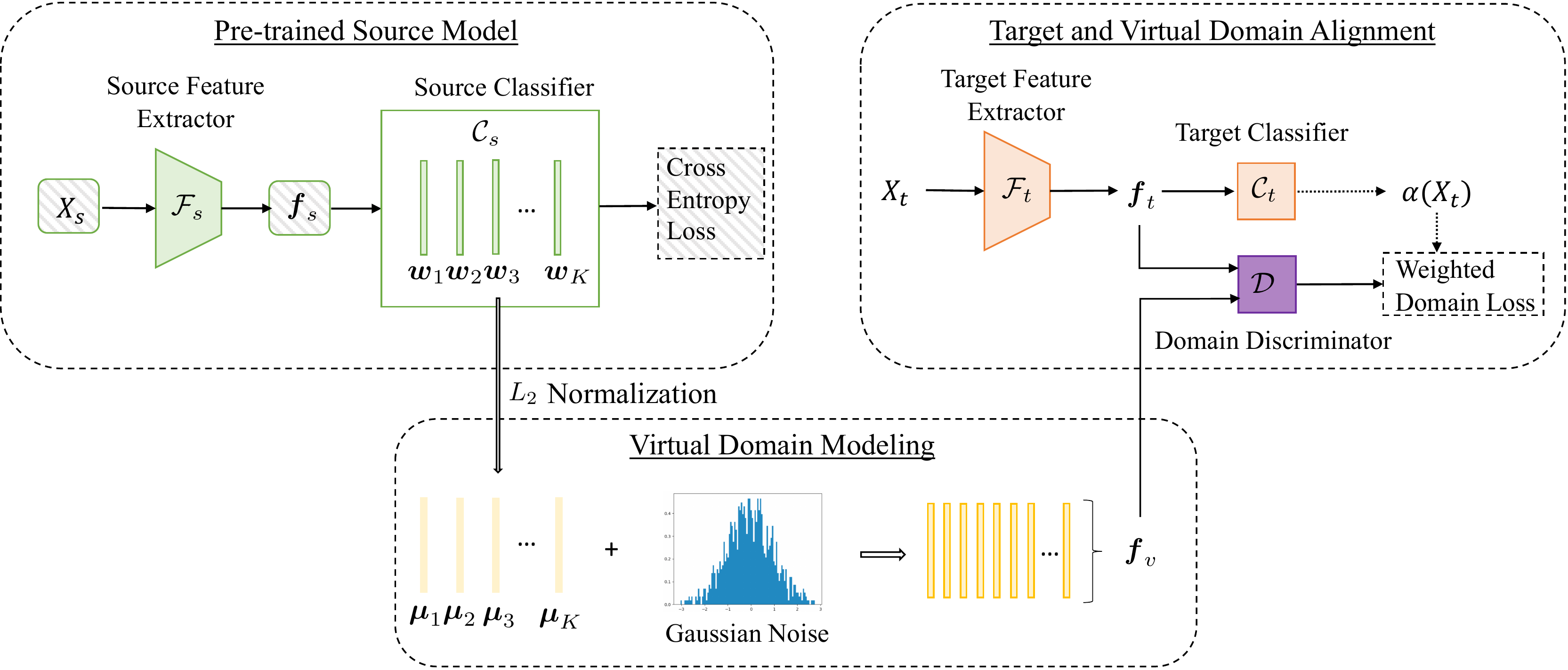}
        \vspace{-10pt}
         \caption{An overview of our proposed VDM-DA method, in which we are only given the pre-trained source model without accessing to the source data under the SFUDA setting. With the pre-trained model parameters, we propose to first model a virtual domain whose data distribution is similar to that of the original source domain in the high-level feature space. After sampling the virtual domain data, we then propose an uncertainty-aware domain alignment method to enforce the alignment between the target domain and the virtual domain as well as simultaneously achieve better intra-class compactness.
        }
        \label{overview}
        
        \vspace{-20pt}
        \end{center}
\end{figure*}

\section{Methodology}


The core issue in unsupervised domain adaptation is to reduce the distribution gap between the source domain and target domain when learning the classification model. A typical deep learning-based domain adaptation model~\cite{Long2015,Long2017,Ganin2016,Bousmalis2017,Tzeng2017,Hoffman2018,Saito2018,Zhang2020} can generally be decomposed into three parts, a classifier module $\mathcal{C}$, a distribution alignment module $\cD$, and a feature extractor module $\mathcal{F}$. In the prevailing paradigm of unsupervised domain adaptation (UDA), one common approach aims at aligning the distributions between the source and target domains in the shared feature space via different statistical measurements, while preserving discriminative information in the source domain. We denote the labeled source domain as $\{(X_s,Y_s)\}$ and the unlabeled target domain as $\{X_t\}$, where $X$ and $Y$ denote the data and labels, respectively. The typical objective function of UDA for classification can be written as follows,
\begin{equation}
    \mathcal{L}_{\text{UDA}}=\mathcal{L}_{cls}(\mathcal{C}(\mathcal{F}(X_s)), Y_s)+\mathcal{L}_{div}(\mathcal{F}(X_s), \mathcal{F}(X_t))
\end{equation}
where $\mathcal{L}_{cls}$ is the classification loss (e.g. the cross-entropy loss) for the source domain labeled data and $\mathcal{L}_{div}$ measures the distribution gap (e.g. MMD and adversarial loss) for modeling the data distribution mismatch between the source domain and the target domain. 

However, for the source data-free domain adaptation (SFUDA), we are only given the pre-trained source model ($\mathcal{F}_s$ and $\mathcal{C}_s$) and unlabeled target domain data ($X_t$) and we cannot access to the source domain data ($X_s$). Hence, we propose to model a virtual domain, such that the domain gap between the source domain and the virtual domain, as well as that between the virtual domain and the target domain can be minimized at the same time. Thus, our newly proposed objective function for SFUDA becomes
\begin{equation}
\begin{split}
    \hspace{-1em}\mathcal{L}_{\text{SFUDA}}=&\mathcal{L}_{cls}(\mathcal{C}(\mathcal{F}(X_s)), Y_s)+\mathcal{L}_{div}(\mathcal{F}(X_s), \mathcal{F}(X_v))\\
    &+\mathcal{L}_{div}(\mathcal{F}(X_v), \mathcal{F}(X_t))
\end{split}   
\end{equation}

The key issue in the above objective function is how to model the virtual domain by minimizing $\mathcal{L}_{div}(\mathcal{F}(X_s), \mathcal{F}(X_v))$ merely based on $\mathcal{F}_s$ and $\mathcal{C}_s$ in the absence of $X_s$. Once the virtual domain is constructed, $\mathcal{L}_{div}(\mathcal{F}(X_v), \mathcal{F}(X_t))$ can be minimized via any distribution divergence loss.

\subsection{Virtual Domain Modeling}
Due to the absence of source data in the source data-free setting, we propose to model a virtual domain to mimic the real source domain data. A typical classification model transforms the original high-dimensional visual data into a compact feature space, where it is much easier to compare and model the data distribution. Thus, it is unnecessary to model the distribution in the original data space. Unlike the Model Adaptation~\cite{Li2020} method that synthesizes the source data in the original data space, we propose to model the virtual domain in a low-dimensional feature space based on the observation that most successful domain adaptation methods for the high-level semantics related tasks (e.g. the classification task) reduce the distribution gap in a more compact feature space rather than in the original data space. 

\subsubsection{GMM-based Virtual Domain Modeling}
Here, we propose to model the virtual domain by using the Gaussian Mixture Model (GMM). Our motivation is that, the features of samples in the high-level feature space learned by using the deep learning methods can be treated as following the mixture of Gaussian distribution for the semantics related tasks such as classification. For example, the data in each class can be assumed to be sampled from one of the Gaussians in the mixture model. As a probabilistic model, GMM assumes all data points are generated from a mixture of a few Gaussian distributions with unknown parameters. Theoretically, the GMM can well approximate any continuous distribution by using a finite number of mixture of Gaussian distributions~\cite{Lindsay1995,McLachlan2019}. Theoretically, a Gaussian mixture model can represent any distribution as follows,
\begin{equation}
\vspace{-0.8em}
    p(\bm{x})=\sum_{k=1}^K \pi_{k} \mathcal{N}(\bm{x}|\bm{\mu}_{k}, \Sigma_{k})
\end{equation}
with $\pi_{k}$ the mixing coefficients, where $\sum_{k=1}^K \pi_{k}=1$ and $\pi_{k}\geq 0$ for any $k$. Hence, four unknown parameters are required to be estimated: 1) the number of Gaussians $K$, 2) the mixing coefficients $\pi_{k}$, 3) the mean of each Gaussian $\bm{\mu}_{k}$, and 4) the covariance of each Gaussian  $\Sigma_{k}$.

In the context of source data-free domain adaptation, the modeled virtual domain is expected to be well aligned with the source domain, such that the target domain can be further aligned with the virtual domain to achieve effective domain adaptation. Thus, we propose to estimate the unknown parameters of the GMM model for the virtual domain such that the distribution gap between the virtual domain and the source domain is minimized. 

The unknown parameters in GMM are typically optimized via some iterative methods, such as the Expectation-Maximization (EM) algorithm. However, since we do not have access to the source domain data, it is infeasible to estimate the parameters in such an iterative way. Here, we approximate these parameters based on our observations and in-depth analysis. 

First, the number of Gaussian distributions $K$ can be naturally set as the number of classes in the source domain. Secondly, the mixing coefficients $\pi_{k}$ are set as $\frac{1}{K}$ by assuming a class-balanced source dataset, which also satisfies the constraints that $\sum_{k=1}^K \pi_{k}=1$ and $\pi_{k}\geq 0$ for any $k$.

Then the key and most challenging task turns to the estimation of $\bm{\mu}_{k}$ and $\Sigma_{k}$ of the GMM model. In the source data-free setting, the underlying distribution cannot be explicitly observed. Thanks to the pre-trained source model, we discover the information hidden in the source model parameters to approximate $\bm{\mu}_{k}$ and $\Sigma_{k}$ of the virtual domain. 

\subsubsection{Approximation of $\bm{\mu}_{k}$ and $\Sigma_{k}$}
\label{sec:para_est}
In a typical deep learning-based architecture for classification, the model generally can be decomposed into a feature extractor module $\mathcal{F}$ and a classifier module $\mathcal{C}$. The feature extractor $\mathcal{F}$ consists of all the layers except for the last fully connected (fc) layer, while the classifier module $\mathcal{C}$ is defined as the last fc layer. After feeding the original input data (e.g. 2D images or 3D point clouds) $X$ into the feature extractor, the data can be represented as a feature vector $\bm{f}=\mathcal{F}(X)$ with more compact semantic information. The classifier module is generally a linear classifier with the output dimension as the number of classes in the current task of-interest. According to one of the most common interpretations, each row of the learned linear classifier's weights corresponds to a template or a prototype for one of the classes~\cite{Chen2019b}. Motivated by this interpretation, we can easily produce the prototypes of all classes via the learned weights of each classifier in the pre-trained source domain model. Therefore, we propose to approximate $\bm{\mu}_{k}$ by defining it as one of the rows of the $L_2$ normalized weights $[\bm{w}_1 , \bm{w}_2 , ..., \bm{w}_K]^T$ from the source classifier.

If the data can be observed in the source domain, after generating the approximated $\bm{\mu}_{k}$, we can directly estimate $\Sigma_{k}$ based on the maximum likelihood approach. However, the key obstacle is still the unavailability of the source data. Hence, by noting that the final goal of our task is still classification, as long as the features sampled from the virtual domain exhibit satisfactory discriminative capability and certain variations, the constructed virtual domain is expected to be sufficient for domain adaptation. For simplicity, we assume that each Gaussian is isotropic and different Gaussian distributions share the same variance in the mixture model. Thus, only one scalar variance parameter $\sigma^2$ needs to be determined in our model. Here, we propose to determine the value of $\sigma^2$ based on the minimum pairwise distance among different prototypes to preserve the discriminative capability and certain variations.
\begin{equation}
\vspace{-0.5em}
  {\sigma}^2 = \frac{1}{\lambda} \min_{\forall m\neq n} {Dist} (\bm{\mu}_m, \bm{\mu}_n).
\label{eq:sigma}
\end{equation}
where $Dist$ is the given distance metric (e.g. Euclidean distance or Cosine similarity), $m,n \in \{1, ..., K\}$, and $\lambda$ is a hyperparameter for controlling the scale of the variance. Intuitively, we need to properly set a value for $\lambda$ to make sure the samples that are close to the decision boundary between the two most confusing classes can still be well separated by the ideal classifier.

\subsubsection{Final Virtual Domain}
Since all the unknown GMM parameters are produced, we can write the distribution of the virtual domain in the feature space as follows,
\begin{equation}
    P_v(\bm{f}_v)= \sum_{k=1}^K \pi_{k} \mathcal{N}(\bm{f}_v|\bm{\mu}_{k}, \sigma^2 I)
\label{eq:virtual}
\end{equation}
where $\bm{f}_v$ represents the virtual domain features, and $\pi_{k}=\frac{1}{K}$.
With the GMM of the virtual domain, we can simply sample data from the model to construct our virtual domain. This can be simply implemented by adding a noise vector sampled based on the Gaussian distribution $\mathcal{N}(\bm{\epsilon}|0, \sigma^2)$ to the mean vector $\bm{\mu}_{k}$ to generate the  feature of one sample from class $k$. Since the virtual domain GMM is constructed and approximated based on the pre-trained model from the source domain, we consider the virtual domain and the source domain are well aligned, at least based on the first-order moment of the class-conditional distributions.

\subsection{Target and Virtual Domain Alignment}
\label{sec:align}
With the constructed virtual domain that well mimics the original source domain distribution in the feature space, we are ready to further align the target domain with the virtual domain. Here, many statistical methods can be readily used, such as Maximum Mean Discrepancy (MMD), Jensen-Shannon (JS) divergence, and Wasserstein distance. Motivated by a simple and effective Adversarial Discriminative Domain Adaptation (ADDA) method~\cite{Tzeng2017}, we choose a simple adversarial training-based strategy, which is equivalent to reducing the JS-divergence between different domains in the feature space.

Specifically, we initialize the target feature extractor $\mathcal{F}_t$ by using the pre-trained source feature extractor $\mathcal{F}_s$. The generated the virtual domain features $\bm{f}_v$ and the extracted target features $\bm{f}_t=\mathcal{F}_t(X_t)$ are then fed into a domain discriminator $\mathcal{D}$. The domain discriminator $\mathcal{D}$ attempts to distinguish the virtual domain features and the target domain features through a binary classification loss with the ground-truth domain labels of source samples and target samples as 1 and 0, respectively. Therefore, our newly proposed objective to learn both $\mathcal{F}_t$ and $\mathcal{D}$ is then formulated as follows,
\begin{equation}
\begin{split}
   \mathcal{L}_{div}=\min_{\mathcal{F}_t}&\max_{\mathcal{D}} \mathbb{E}_{\bm{f}_v \sim P_v} \log \mathcal{D} (\bm{f}_v) \\
   &+ \mathbb{E}_{X_t \sim P_t} \log(1-\mathcal{D}(\mathcal{F}_t(X_t))),    
\end{split}
\label{eq:vanilla}
\end{equation}
where $P_v$ and $P_t$ are the distributions of the virtual domain in the high-level feature space and the target domain in the original data space, respectively.

\subsubsection{Enhancing Target Domain Compactness}
\label{sec:compact}
Since the variance of the virtual domain is controllable, and a relatively small variance may lead to a more class-concentrated domain, which is helpful for the classification tasks in both the virtual domain and the target domain. However, the real target domain data may exhibit a larger variation, so direct alignment between the virtual domain and the target domain may not be sufficient. Here, to further enhance the intra-class \textit{compactness} for the target domain data, we propose a new mechanism to enforce the uncertain target samples that lie around the decision boundary to be more confident. 

We observe that the uncertain target samples with low prediction confidence scores generally lead to poor intra-class compactness. Thus, we further align the target uncertain target samples with the virtual domain samples by using a newly proposed re-weighting mechanism. Specifically, the more uncertain (\textit{resp.}, certain) the target sample is, a larger (\textit{resp.}, smaller) weight will be assigned when calculating the distribution divergence loss. Formally, the weights that characterize the uncertainty are defined based on a normalized entropy as follows,
\begin{equation}
   \alpha(X_t) = \frac{-\sum_{k=1}^K\delta_k(\mathcal{C}_t(\mathcal{F}_t(X_t)))log\delta_k(\mathcal{C}_t(\mathcal{F}_t(X_t)))}{-\sum_{k=1}^K\frac{1}{K}log \frac{1}{K}},
\label{eq:target_ent}
\end{equation}

where $\delta_k(\mathcal{C}_t(\mathcal{F}_t(X_t)))$ denotes the $k$-th element of the softmax output. 
The distribution divergence loss then becomes,
\begin{equation}
\begin{split}
   \mathcal{L}_{div}=\min_{\mathcal{F}_t}&\max_{\mathcal{D}}\mathbb{E}_{\bm{f}_v \sim P_v} \log \mathcal{D} (\bm{f}_v) \\
   &+ \mathbb{E}_{X_t \sim P_t} \alpha(X_t)\log(1-\mathcal{D}(\mathcal{F}_t(X_t)))
\label{eq:tc_loss}
\end{split}  
\end{equation}

However, one may argue that if a larger domain loss is applied to the more uncertain samples, they might be aligned to a wrong class since these samples are the most confusing ones. We tackle this problem by refining the target model based on the more certain target samples with their assigned pseudo labels. The decision boundary will thus be adapted more towards the target domain data to alleviate the issues of aligning uncertain target samples to a wrong side of the decision boundary. In details, we rank the target samples based on their entropy values in a descending order and choose the first $r\%$ samples as the most certain target samples. These samples are assigned with the pseudo labels predicted by the aligned model in order to further refine the target model.

\subsection{Theoretical Analysis}
In this section, we analyse our method from the theoretical perspective. The seminal works~\cite{Ben-David2010,Mansour2009} on theoretical analysis for domain adaptation have provided the generalization bound of the expected classification error $\epsilon_{P_t}(h)$ in the target domain based on the source domain classification error $\epsilon_{P_s}(h)$ and the domain discrepancy. Formally, let us denote any hypothesis by $h\in \mathcal{H}$, the generalization bound is defined as
\begin{equation}
    \epsilon_{P_t}(h) \leq \epsilon_{P_s}(h)+d(P_s, P_t) + \gamma
\label{eq:theory}
\end{equation}
where $d(P_s, P_t)$ is the domain discrepancy, and $\gamma$ is a constant term. It is crucial to reduce the domain discrepancy and achieve a tighter bound, which have been implemented in different ways, such as the popular  $\mathcal{H}\Delta\mathcal{H}$-Divergence\cite{Ben-David2010,Mansour2009}, Maximum Mean Discrepancy (MMD)~\cite{Pan2009,Long2015},  JS-divergence ~\cite{Ganin2016,Tzeng2017} or Wasserstein distance~\cite{Shen2018}, etc.

In the source data-free setting, we model a virtual domain distribution $P_v$ to mimic the source distribution $P_s$. To reduce the classification error on the target domain, the distributions $P_s$, $P_v$, and $P_t$ should be substantially similar to each other. Hence, the generalization bound in our method becomes
\vspace{-0.5em}
\begin{equation}
    \epsilon_{P_t}(h) \leq  \epsilon_{P_v}(h)+d(P_v, P_t) + \gamma_1,
\label{eq:SFtheory}
\end{equation}
\vspace{-0.5em}
where 
\begin{equation}
    \epsilon_{P_v}(h) \leq  \epsilon_{P_s}(h)+d(P_v, P_s) + \gamma_2
\label{eq:SFtheory}
\end{equation}
Thus, the final bound is defined as,
\begin{equation}
    \epsilon_{P_t}(h) \leq  \epsilon_{P_s}(h)+d(P_s, P_v) +d(P_v, P_t) + \hat{\gamma}
\label{eq:SFtheory}
\end{equation}
where $\hat{\gamma} = \gamma_1+\gamma_2$ is the constant term. 

Instead of reducing $d(P_s, P_t)$ in (\ref{eq:theory}), our goal is to reduce both $d(P_s, P_v)$ and $d(P_v, P_t)$ Here, we analyze the two terms one-by-one. First, in our formula of modeling the virtual domain in Eq.~(\ref{eq:virtual}), the distribution divergence in each class between the source domain and the virtual domain is minimized in terms of the first-order moments. The justification is that the class mean $\bm{\mu}_k$ is defined as the learned weights of the source classifier and these weights can been explained as the means (\ie, the prototypes) of the source classes. Thus, the distribution discrepancy $d(P_s, P_v)$ can be treated as the MMD on the samples in each class from different domains. Secondly, through a domain discriminator, the distribution discrepancy $d(P_v, P_t)$ is approximated by JS-divergence~\cite{Goodfellow2014}, which shares the similar spirit from several existing adversarial learning-based domain adaptation methods~\cite{Ganin2016,Tzeng2017}. In summary, the newly proposed VDM-DA method for SFUDA can still achieve a tight generalization bound for reasonably minimizing the target domain classification error.

\section{Experiments}
In this section, to verify the effectiveness of our proposed VDM-DA method, we conduct extensive experiments on three benchmark datasets for cross-domain object recognition based on both 2D images and 3D point clouds.

\subsection{Experimental Setup}
\noindent\textbf{Datasets.}
Two commonly used 2D image benchmarks (Office31 and VisDA17) and one recently released 3D point-cloud dataset (PointDA-10) are used for evaluation. 
\begin{enumerate}
   \item \textit{Office31} contains three domains (Amazon (A), DSLR (D), and Webcam (W)) and each domain consists of 31 object classes in the office environment. There are in total 4110 images in this dataset. We perform domain adaptation by using each pair of domains, which leads to 6 different tasks.
   \vspace{-8pt}
   \item \textit{VisDA17} is a challenging large-scale benchmark datasets, which includes the images from 12 object classes. It contains two domains(\ie, the synthesis image domain and the real image domain) and the goal is to perform domain adaptation from the synthesis image domain (\ie, the source domain) to the real image domain (\ie, the target domain). The source domain contains 152,409 of synthetic images generated by rendering 3D models, while the target domain has 55,400 real images sampled from Microsoft COCO.
    \vspace{-8pt}
   \item \textit{PointDA-10} dataset is a 3D point cloud domain adaptation benchmark used in~\cite{Qin2019}, which contains 3D point clouds of 10 classes from three domains, ModelNet40 (Mo), ShapeNet (Sh) and Scannet (Sc). Each domain contains its own training and testing sets. We perform domain adaptation by using the training set of one domain as the labeled source domain, the training set of another domain as the unlabeled target domain and further evaluate on the testing set of this unlabeled target domain, which leads to 6 tasks.
\end{enumerate}

\noindent \textbf{Network Architecture.} 
For fair comparison, we use the same backbone networks as in the previous domain adaptation methods~\cite{Liang2020}. As described in Section~\ref{sec:method}, a typical domain adaptation model is composed of a feature extractor $\mathcal{F}$ (\eg, $\mathcal{F}_s$ / $\mathcal{F}_t$), a classifier $\mathcal{C}$ (\eg, $\mathcal{C}_s$ / $\mathcal{C}_t$), and a domain discriminator $\mathcal{D}$. For the 2D image recognition task, the pre-trained ResNet-50 or ResNet-101~\cite{He2016} with one additional
bottleneck $\textit{fc}$ layer (with 256 units) is used as the feature extractor. For the 3D point cloud recognition task, as suggested in~\cite{Qin2019}, pre-trained PointNet~\cite{Qi2017} with two additional
\textit{fc} layers are used. For both tasks, the classifier $\mathcal{C}$ is defined as the last
\textit{fc} layer, while the domain discriminator $\mathcal{D}$ consists of three \textit{fc} layers
(\ie, $\bm{f}_t/\bm{f}_v$ $\rightarrow 1024\rightarrow 1024\rightarrow 2$).

\noindent \textbf{Implementation Details.}
We optimize the whole network parameters  by using the SGD optimizer with momentum of $0.9$, weight decay of $1e^{-3}$ and the batch size of 32 on all datasets. The initial learning rate $\eta_0$ is empirically set to $1e^{-2}$, then the learning rate after each iteration is decreased based on the equation $\eta=\eta_0\cdot(1+10\cdot p)^{-0.75}$ as suggested in~\cite{Ganin2016}, where $p$ changes from 0 to 1 during the training process. During training, feature extractor $\mathcal{F}$ is learned with the learning rate as 0.1 times that of the current step. We empirically set $r\%=70\%$ for the 2D image and $r\%=30\%$ for the point cloud related experiments.

\noindent\textbf{Baseline Methods.}
We compare our method VDM-DA with both traditional UDA methods and source data-free UDA (SFUDA) methods. The UDA baselines include  DANN~\cite{Ganin2016}, ADDA~\cite{Tzeng2017}, DAN~\cite{Long2015}, JAN~\cite{Long2017}, CDAN~\cite{Long2018}, MCD~\cite{Saito2018}, BSP~\cite{Chen2019a}, SAFN~\cite{Xu2019} and STAR~\cite{Lu2020}. While the SFUDA baseline methods are SHOT~\cite{Liang2020}, BAIT~\cite{Yang2020}, and ModelAdapt~\cite{Li2020}. For point cloud related tasks, we additionally compare our method with PointDAN~\cite{Qin2019}, which is the state-of-the-art domain adaptation method for 3D point clouds. The results of the baseline methods are copied from their original works.

\begin{table*}[!htb]
\begin{footnotesize}
\begin{center}
\vspace{-1em}
\caption{Accuracies (\%) of different cross-domain object recognition methods on the VisDA17 benchmark. }
\vspace{-8pt}
\label{tab:visda}
\begin{tabular}{llllllllllllll}
\hline
Methods & plane & bcycle & bus & car & horse & knife & mcycle & person & plant & sktbrd & train & truck & Avg. \\
\hline \hline
ResNet101~\cite{He2016} & 55.1 & 53.3 & 61.9 & 59.1 & 80.6 & 17.9 & 79.7 & 31.2 & 81.0 & 26.5 & 73.5 & 8.5 & 52.4 \\
DANN~\cite{Ganin2016} & 81.9 & 77.7 & 82.8 & 44.3 & 81.2 & 29.5 & 65.1 & 28.6 & 51.9 & 54.6 & 82.8 & 7.8 & 57.4 \\
DAN~\cite{Long2015} & 87.1 & 63.0 & 76.5 & 42.0 & 90.3 & 42.9 & 85.9 & 53.1 & 49.7 & 36.3 & 85.8 & 20.7 & 61.1 \\
CDAN~\cite{Long2018} & 85.2 & 66.9 & 83.0 & 50.8 & 84.2 & 74.9 & 88.1 & 74.5 & 83.4 & 76.0 & 81.9 & 38.0 & 73.9 \\
MCD~\cite{Saito2018} & 87.0 & 60.9 & 83.7 & 64.0 & 88.9 & 79.6 & 84.7 & 76.9 & 88.6 & 40.3 & 83.0 & 25.8 & 71.9 \\
BSP~\cite{Chen2019a}  & 92.4 & 61.0 & 81.0 & 57.5 & 89.0 & 80.6 & 90.1 & 77.0 & 84.2 & 77.9 & 82.1 & 38.4 & 75.9 \\
SAFN~\cite{Xu2019} & 93.6 & 61.3 & 84.1 & 70.6 & 94.1 & 79.0 & \textbf{91.8} & 79.6 & 89.9 & 55.6 & 89.0 & 24.4 & 76.1 \\
STAR~\cite{Lu2020} & 95.0 & 84.0 & \textbf{84.6} & 73.0 & 91.6 & 91.8 & 85.9 & 78.4 & 94.4 & 84.7 & 87.0 & 42.2 & 82.7 \\
\hline
SHOT~\cite{Liang2020} & 94.3 & 88.5 & 80.1 & 57.3 & 93.1 & 94.9 & 80.7 & 80.3 & 91.5 & \textbf{89.1} & 86.3 & \textbf{58.2} & 82.9 \\
ModelAdapt~\cite{Li2020} & 94.8 & 73.4 & 68.8 & \textbf{74.8} & 93.1 & 95.4 & 88.6 & \textbf{84.7} & 89.1 & 84.7 & 83.5 & 48.1 & 81.6 \\
BAIT~\cite{Yang2020} & 93.7 & 83.2 & 84.5 & 65.0 & 92.9 & 95.4 & 88.1 & 80.8 & 90.0 & 89.0 & 84.0 & 45.3 & 82.7 \\
VDM-DA (w/o TC) & 95.1 & 88.3 & 77.4 & 61.5 & 93.6 & 95.2 & 81.0 & 80.9 & 94.1 & 80.2 & 86.6 & 55.9 & 82.5 \\
VDM-DA & \textbf{96.9} & \textbf{89.1} & 79.1 & 66.5 & \textbf{95.7} & \textbf{96.8} & 85.4 & 83.3 & \textbf{96.0} & 86.6 & \textbf{89.5} & 56.3 & \textbf{85.1}
\\
\hline
\end{tabular}
\end{center}
\end{footnotesize}
\vspace{-15pt}
\end{table*}

\begin{table}[!h]
\begin{footnotesize}
\begin{center}
\vspace{-1em}
\caption{Accuracies (\%) of different cross-domain object recognition methods on the Office-31 benchmark. }
\vspace{-8pt}
\label{tab:office31}
\begin{tabular}{m{2.3cm}m{0.43cm}m{0.43cm}m{0.43cm}m{0.43cm}m{0.43cm}m{0.43cm}m{0.4cm}}
\hline
Methods & A→W & A→D & W→A & W→D & D→A & D→W & Avg. \\
\hline
\hline 
ResNet50~\cite{He2016} & 75.8 & 79.3 & 63.8 & 99.0 & 63.6 & 95.5 & 79.5 \\
DANN~\cite{Ganin2016} & 82.0 & 79.7 & 67.4 & 99.1 & 68.2 & 96.9 & 82.2 \\
DAN~\cite{Long2015} & 84.2 & 87.3 & 65.2 & \textbf{100.0} & 66.9 & 98.4 & 83.7 \\
JAN~\cite{Long2017} & 93.7 & 89.4 & 71.0 & \textbf{100.0} & 71.2 & 98.4 & 87.3 \\
CDAN~\cite{Long2018} & 94.1 & 92.9 & 69.3 & \textbf{100.0} & 71.0 & 98.6 & 87.7 \\
MCD~\cite{Saito2018} & 88.6 & 92.2 & 69.7 & \textbf{100.0} & 69.5 & 98.5 & 86.5 \\
BSP~\cite{Chen2019a} & 93.3 & 93.0 & 72.6 & \textbf{100.0} & 73.6 & 98.2 & 88.5 \\
SAFN~\cite{Xu2019} & 90.3 & 92.1 & 71.2 & \textbf{100.0} & 73.4 & \textbf{98.7} & 87.6 \\
STAR~\cite{Lu2020} & 92.6 & 93.2 & 70.0 & \textbf{100.0} & 71.4 & \textbf{98.7} & 87.8 \\
\hline
SHOT~\cite{Liang2020} & 90.1 & \textbf{94.0} & 74.3 & 99.9 & 74.7 & 98.4 & 88.6 \\
ModelAdapt~\cite{Li2020} & 93.7 & 92.7 & \textbf{77.8} & 99.8 & 75.3 & 98.5 & 89.6 \\
BAIT~\cite{Yang2020} & \textbf{94.6} & 92.0 & 75.2 & \textbf{100.0} & 74.6 & 98.1 & 89.1\\
VDM-DA   (w/o TC) & 94.0 & 93.0 & 76.6 & \textbf{100.0} & 75.4 & 98.0 & 89.5 \\
VDM-DA & 94.1 & 93.2 & 77.1 & \textbf{100.0} & \textbf{75.8} & 98.0 & \textbf{89.7}\\
\hline
\end{tabular}
\end{center}
\end{footnotesize}
\vspace{-20pt}
\end{table}

\subsection{Experimental Results}
\subsubsection{2D Image Recognition}

Table~\ref{tab:visda} and~\ref{tab:office31} show the experimental results on two 2D image-based cross-domain object recognition benchmarks. ResNet50 and ResNet101 are used as the backbone network in all methods for Office-31 and VisDA17 benchmarks, respectively.
We also report the results of a variant of our method without enhancing target domain intra-class compactness (denoted by ``VDM-DA (w/o TC)''), in which $\alpha(X_t)$ in Eq.~(\ref{eq:target_ent}) is set as uniform weights for all target data. 

From the results, we observe that our proposed VDM method outperforms both state-of-the-art UDA and SFUDA methods on both datasets. In particular, when compared to the traditional UDA methods which use the real source data for reducing domain distribution mismatch, our VDM-DA approach is able to control the variance of virtual domain samples. Consequently, the generated samples are more class-balanced, compact, and discriminative with less noisy outliers, which leads to better results. Moreover, our VDM-DA method also outperforms the recent SFUDA methods SHOT~\cite{Liang2020}, BAIT~\cite{Yang2020} and ModelAdapt~\cite{Li2020}, as we explicitly reduce the distribution divergence between source and target domains. 

Moreover, without enhancing the intra-class compactness in the target domain, we observe that the results of the alternative method VDM-DA(w/o TC) are worse than our VDM-DA on the two benchmarks, which verifies that it is beneficial to improve the intra-class compactness in the target domain.

\vspace{-1em}
\subsubsection{3D Point Cloud Recognition}
\vspace{-1em}
\begin{table}[!h]
\begin{footnotesize}
\begin{center}
\caption{Accuracyies (\%) of different cross-domain object recognition methods on the PointDA-10 benchmark. }
\vspace{-8pt}
\label{tab:3D}
\begin{tabular}{m{2.3cm}m{0.43cm}m{0.43cm}m{0.43cm}m{0.43cm}m{0.43cm}m{0.43cm}m{0.4cm}}
\hline
Methods & mo-sh & mo-sc & sh-mo & sh-sc & sc-mo & sc-sh & Avg. \\
\hline \hline
PointNet~\cite{Qi2017} & 42.5 & 22.3 & 39.9 & 23.5 & 34.2 & 46.9 & 34.9 \\
MMD~\cite{Long2013} & 57.5 & 27.9 & 40.7 & 26.7 & 47.3 & 54.8 & 42.5 \\
DANN~\cite{Ganin2016} & 58.7 & 29.4 & 42.3 & 30.5 & 48.1 & 56.7 & 44.2 \\
ADDA~\cite{Tzeng2017} & 61.0 & 30.5 & 40.4 & 29.3 & 48.9 & 51.1 & 43.5 \\
MCD~\cite{Saito2018} & 62.0 & 31.0 & 41.4 & 31.3 & 46.8 & 59.3 & 45.3 \\
PointDAN~\cite{Qin2019} & \textbf{64.2} & \textbf{33.0} & 47.6 & 33.9 & \textbf{49.1} & \textbf{64.1} & 48.7 \\
\hline 
SHOT~\cite{Liang2020} & 55.1 & 23.8 & 56.2 & 29.6 & 40.2 & 43.6 & 41.4 \\
VDM-DA~(w/o~TC) & 54.9 & 29.1 & 55.1 & 31.4 & 42.1 & 55.3 & 44.6 \\
VDM-DA & 58.4 & 30.9 & \textbf{61.0} & \textbf{40.8} & 45.3 & 61.8 & \textbf{49.7}\\
\hline
\end{tabular}
\end{center}
\end{footnotesize}
\vspace{-8pt}
\end{table}

\begin{figure*}[!h]
\vspace{-0.8em}
    \centering
    \begin{subfigure}[b]{0.2368\textwidth}
     \centering
     \caption{Source and Virtual Domains}
     \vspace{-5pt}
     \includegraphics[width=\textwidth]{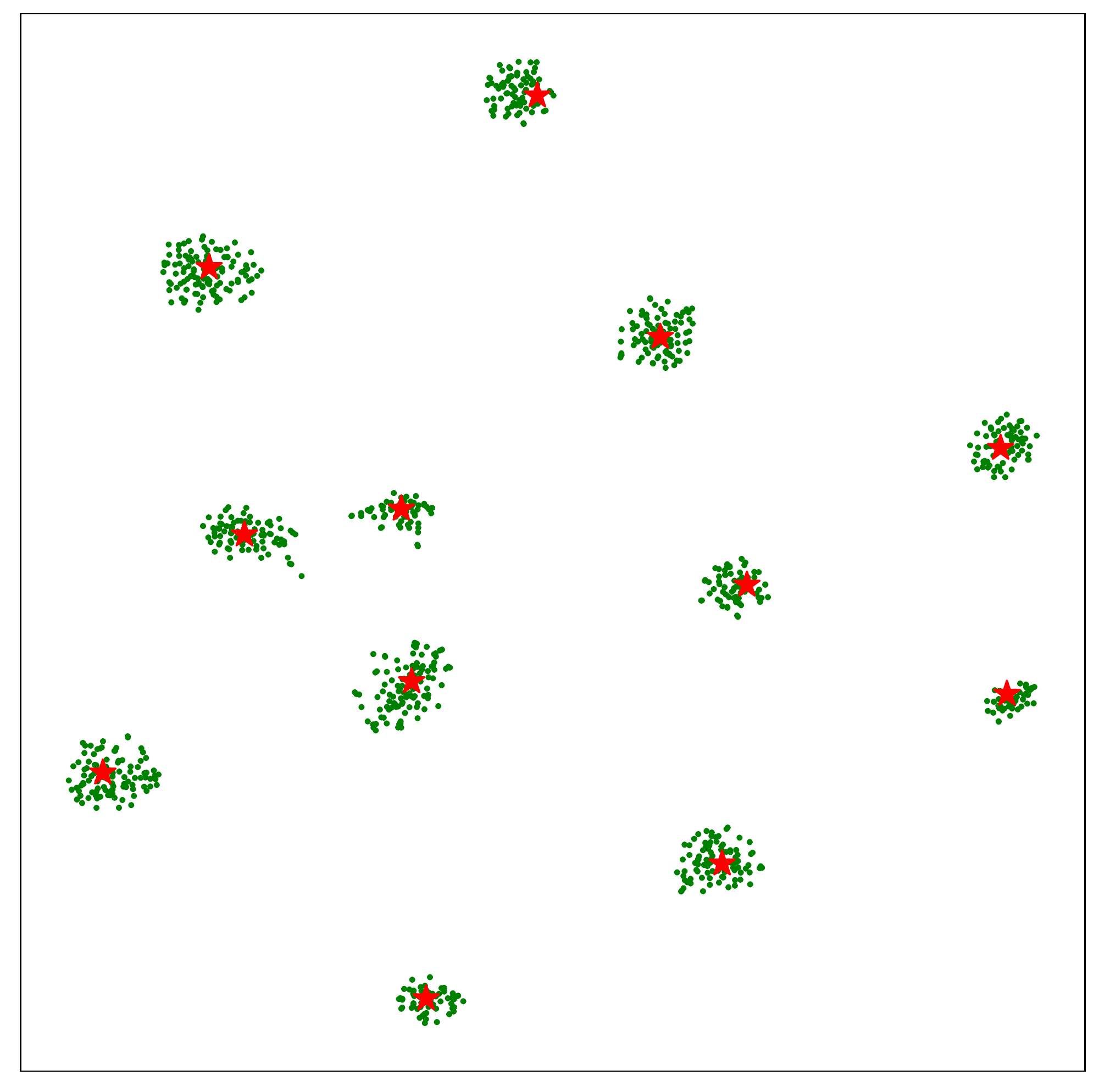}
     \vspace{-9pt}
     \label{fig:tsne_source}
    \end{subfigure}
    \hfill \rulesep
    \begin{subfigure}[b]{0.74\textwidth}
        \centering
        \caption{Target and Virtual Domains}
        \vspace{-5pt}
        \begin{subsubfigure}[b]{0.32\textwidth}
         \centering
         \includegraphics[width=\textwidth]{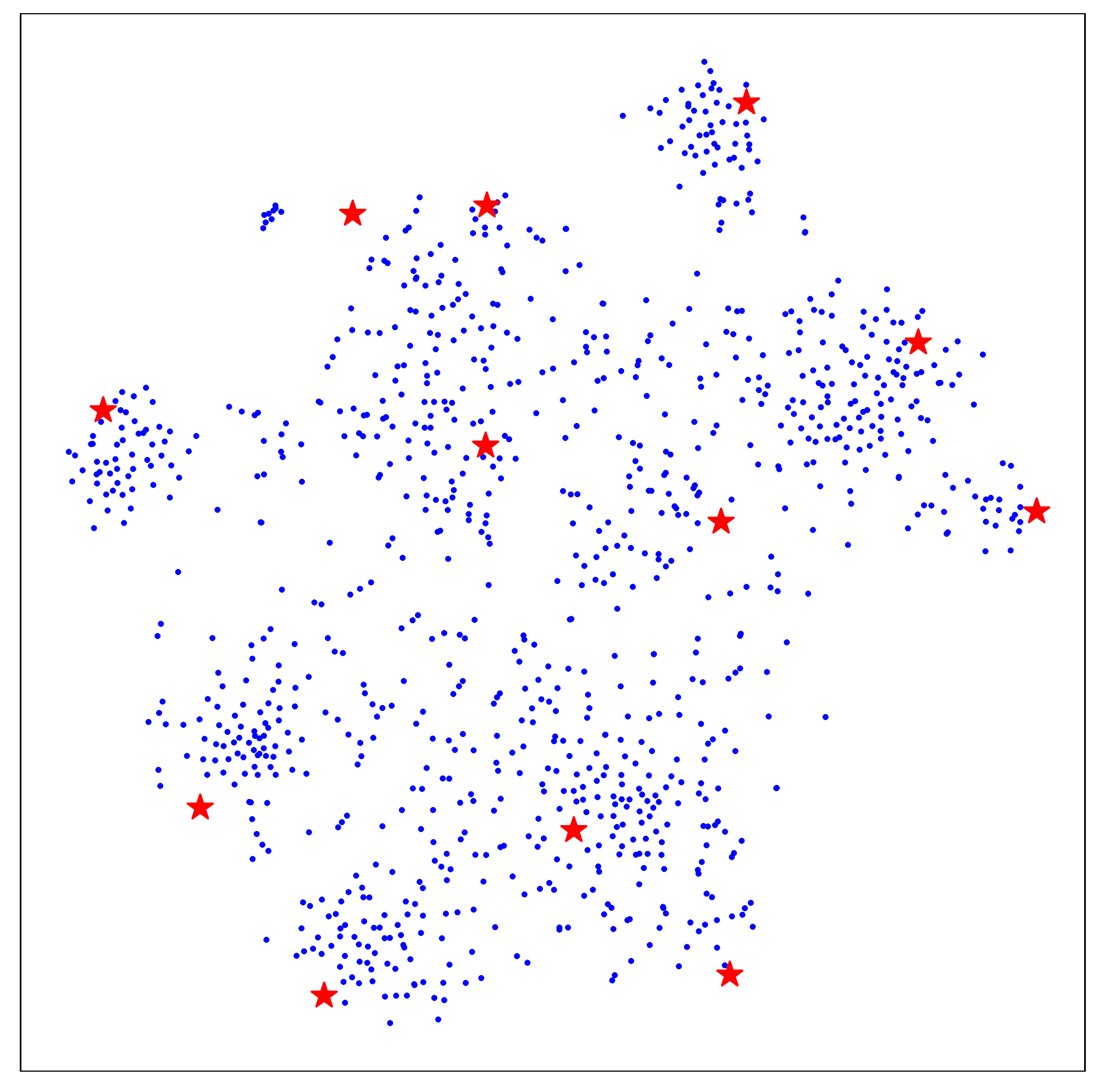}
         \vspace{-15pt}
         \caption{Before Alignment}
         \label{fig:tsne_before}
        \end{subsubfigure}
        \hfill
        \begin{subsubfigure}[b]{0.32\textwidth}
         \centering
         \includegraphics[width=\textwidth]{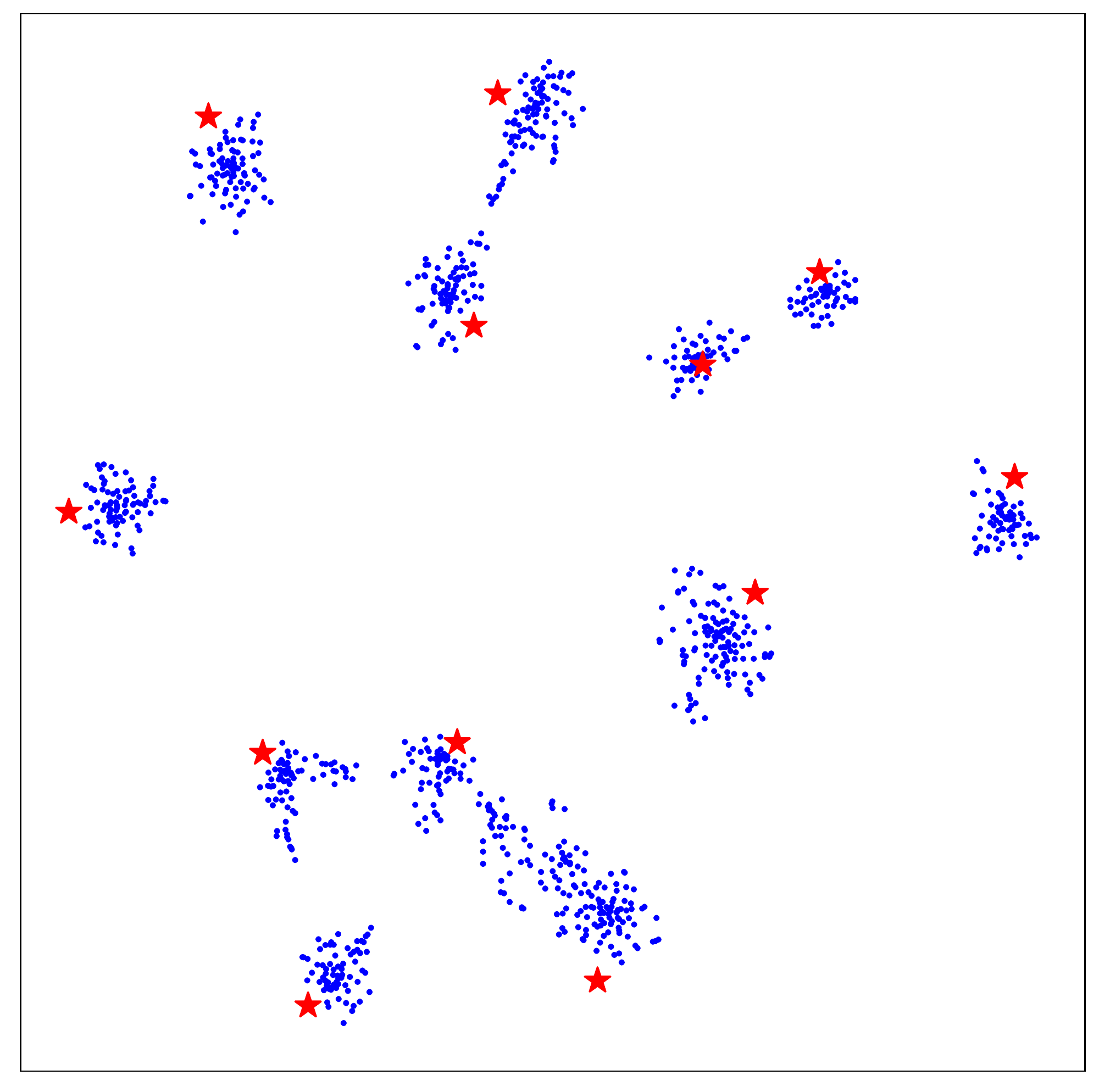}
         \vspace{-15pt}
         \caption{Vanilla Alignment}
         \label{fig:tsne_vanilla}
        \end{subsubfigure}
        \hfill
        \begin{subsubfigure}[b]{0.32\textwidth}
         \centering
         \includegraphics[width=\textwidth]{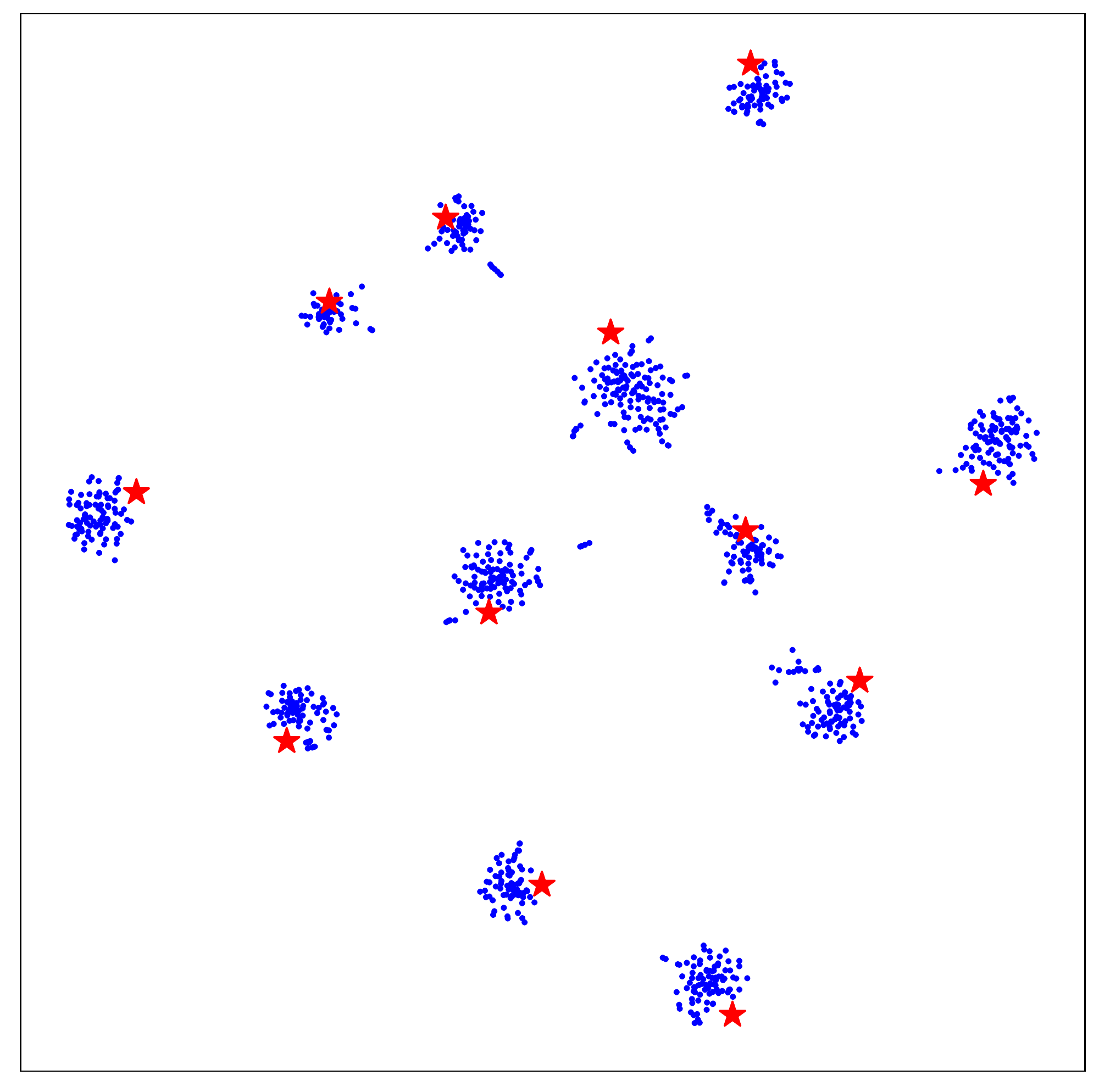}
         \vspace{-15pt}
         \caption{Uncertainty-aware Alignment}
         \label{fig:tsne_uncertain}
        \end{subsubfigure}
     \vspace{-15pt}
    \label{fig:tsne_target}
    \end{subfigure}
    \caption{The t-SNE visualization results on the VisDA17 dataset. (a) the randomly sampled real source data (green dots) and virtual domain data (red stars), where the locations and sizes of the red stars indicate the mean $\bm{\mu}$ and the variance $\sigma^2$ of each class in the virtual domain, respectively. (b) the randomly sampled target data (blue dots) and virtual domain data (red stars) when (i) before alignment, (ii) after alignment by using the vanilla alignment method, and (iii) the newly proposed uncertainty-aware alignment method.
    }
    \label{fig:tsne_vt}
    \vspace{-12pt}
\end{figure*}

In Table~\ref{tab:3D}, we report the results of all methods for the cross-domain 3D point cloud recognition task. Similarly as in the image recognition task, our newly proposed VDM-DA method outperforms all the existing UDA baseline methods in terms of the average accuracy. Specifically, our VDM-DA method achieves better average accuracy than PointDAN, which is specially designed for the cross-domain 3D recognition task with additional complicated modules for both local and global alignment. These results clearly verify that it is beneficial to use the generated virtual domain data from our VDM-DA for different domain adaptation tasks.

We also run the original codes from the work of SHOT on the 3D point cloud datasets. However, the results show that the baseline method SHOT highly relies on the degree of distribution mismatch. In other words, the self-training strategy used in the baseline method SHOT may not work well if the source domain and the target domain have considerable distribution gap. In contrast, our generated virtual domain data shares more similar distribution as the source domain and this can be 
used as a bridge to reduce data distribution mismatch between the source and target domains.

\subsection{Parameter Sensitivity and Qualitative Analysis}
In this section, we take the VisDA17 dataset as an example to conduct more experiments to evaluate the robustness of our method when using different parameters, and also provide qualitative results to further demonstrate the effectiveness of our method. 
\begin{table}[!h]
\begin{small}
\caption{Results of our VDM-DA method when using different $\lambda$ values on the VisDA17 dataset}
\vspace{-10pt}
\label{tab:lambda}
\begin{tabular}{c|cccccc}
\hline
$\lambda$ & 2 & 4 & 6 & 8 & 10 & 12 \\
\hline 
Accuracies(\%) & 83.3 & 84.2 & 85.1 & 84.7 & 84.2 & 82.2 \\
\hline
\end{tabular}
\vspace{-10pt}
\end{small}
\end{table}

\noindent\textbf{Robustness to $\lambda$.} When generating the virtual domain samples, the hyperparameter in Eq.(\ref{eq:sigma}) controls the intra-class variance of the generated virtual domain samples. On one hand, if $\lambda$ is too small, the intra-class variance is large, so it is difficult to distinguish the virtual domain samples from different classes. On the other hand, if $\lambda$ is too large, the virtual domain samples in each class will be concentrated only in a compact area around the class mean, making it more difficult to align the target domain and the virtual domain, as the real samples usually come from a more diverse distribution. Fortunately, the newly proposed strategy for enhancing the intra-class compactness in the target domain (see Section~\ref{sec:compact}) would help us to partially solve this issue, allowing us to choose $\lambda$ in a relatively large range. As shown in Table~\ref{tab:lambda}, we observe that our method can achieve promising results when varying $\lambda$ in the range of $[4, 10]$.

\noindent\textbf{Visualization of Virtual Domain Samples.} A good virtual domain should exhibit two key properties: 1) its data distribution should be similar as the real source domain and 2) the virtual domain samples from different classes are well separated. To verify the two properties, Figure~\ref{fig:tsne_source} visualizes the t-SNE results of the randomly sampled real source data and the data randomly sampled from the virtual domain. The results show that the virtual domain samples are well aligned with the real source data. In addition, since we can manually control the variance during the virtual domain modeling process, the sampled virtual domain data are more compact than the source data and contain much less outlier samples that lie between any two confusion classes.

\noindent\textbf{Visualization of Target Domain Compactness.} As discussed in Section~\ref{sec:align}, the target domain samples can be quite diverse, and improving the intra-class compactness in the target domain would help distribution alignment between the target domain and the virtual domain. To validate this, in Figure~\ref{fig:tsne_target}, we use t-SNE to visualize the sampled from both the target and virtual domains. Three methods are evaluated, the source only method (Figure~\ref{fig:tsne_target}(i)), the vanilla alignment method (\ie by using Eq.~(\ref{eq:vanilla})) without considering uncertainty (Figure~\ref{fig:tsne_target}(ii)), and our newly proposed uncertainty-aware alignment method (Figure~\ref{fig:tsne_target}(iii)). It can be seen that while the vanilla alignment method can align the target samples with the virtual domain class prototypes to a certain degree, our uncertainty-aware alignment method achieves better alignment results and produces more class concentrated representations.

\section{Conclusion}
In this work, we have proposed a new method called Domain Adaptation with Virtual Domain Modeling (VDM-DA) for source data-free unsupervised domain adaptation (SFUDA). To learn domain-invariant representations to reduce distribution gap between the inaccessible source domain and the target domain, we propose to model an intermediate virtual domain which has similar data distribution as the unobserved source data in the high-level feature space by using a Gaussian Mixture Model (GMM),
our virtual domain construction procedure is simple without introducing any extra learnable parameters. We further align the target domain to the virtual domain by using our newly proposed uncertainty-aware alignment strategy to improve the intra-class compactness in the target domain. Extensive experiments on both 2D images and 3D point-clouds for different cross-domain object recognition tasks demonstrate the effectiveness of our proposed VDM-DA method.

{\small
\bibliographystyle{ieee_fullname}
\bibliography{egbib}
}

\end{document}